# Real-world Deployment and Evaluation of PErioperative AI CHatbot (PEACH) - a Large Language Model Chatbot for Perioperative Medicine

Yu He Ke*1,2,3
Liyuan Jin4,5
Kabilan Elangovan5
Bryan Wen Xi Ong6
Chin Yang Oh6
Jacqueline Sim1
Kenny Wei-Tsen Loh1
Chai Rick Soh1,3
Jonathan Ming Hua Cheng1
Aaron Kwang Yang Lee1
Daniel Shu Wei Ting3,4,5
Nan Liu3
Hairil Rizal Abdullah1,2,3

Affiliations:
1 Department of Anesthesiology, Singapore General Hospital, Singapore, Singapore
2 Data Science and Artificial Intelligence Lab, Singapore General Hospital, Singapore, Singapore
3 Duke-NUS Medical School, Singapore, Singapore
4 Singapore National Eye Centre, Singapore Eye Research Institute, Singapore, Singapore
5 Singapore Health Services, Artificial Intelligence Office, Singapore, Singapore
6 Open Government Products, Singapore, Singapore

*Corresponding Author

Corresponding Author:
Name: Ke Yuhe
Email: ke.yuhe@singhealth.com.sg
Address: 31 Third Hospital Ave. Singapore 168753
Institution: Department of Anesthesiology, Singapore General Hospital, Singapore, Singapore

**Figure source:**
All figures included in this study are original and generated solely by the authors.

**Key Points**

**Question:**
Can a secure, large language model-based chatbot be safely and accurately integrated into real-world preoperative clinical decision-making?

**Findings:**
In this real-world evaluation involving 240 clinical scenarios, the PErioperative AI CHatbot (PEACH) achieved an overall accuracy of 97.9% after protocol refinements, with minimal hallucinations (0.4%) and deviations (0.8%). High user acceptability and agreement with clinical decisions were observed.

**Meaning:**
PEACH demonstrated strong potential as a reliable and adaptable tool that improves consistency, accuracy, and efficiency in perioperative decision-making.

**Abstract**

**Background**
Large Language Models (LLMs) are emerging as powerful tools in healthcare, particularly for complex, domain-specific tasks. This study describes the development and evaluation of the PErioperative AI CHatbot (PEACH), a secure LLM-based system integrated with local perioperative guidelines to support preoperative clinical decision-making.

**Methods**
We embedded PEACH with 35 institutional perioperative protocols in the secure Claude 3.5 Sonet LLM framework within Pair Chat (developed by Singapore Government). The system was tested with a silent deployment with real-world data. Accuracy, safety, and usability were assessed. Deviations and hallucinations were categorized based on potential harm, and user feedback was evaluated using the Technology Acceptance Model (TAM). Updates to PEACH was made after the initial silent deployment to make minor amendments to one of the protocol.

**Results**
A total of 240 real-world clinical iterations were evaluated. PEACH achieved a first-generation accuracy of 97.5% (78/80) and an overall accuracy of 96.7% (232/240) across three iterations. The updated PEACH had accuracy improved to 97.9% (235/240), with a statistically significant difference from the null hypothesis of 95% accuracy ($p = 0.018$, 95% CI: 0.952–0.991). Hallucinations and deviations were minimal (both 1/240 and 2/240 respectively). There were high usability, with clinicians noting that PEACH expedited decisions in 95% of cases. The inter-rater reliability was observed within PEACH were (kappa 0.772–0.893) and (0.610–0.784) among attendings.

**Conclusion**
PEACH is an accurate, adaptable tool that enhances consistency and efficiency in perioperative decision-making. Future research should explore its scalability across specialties and its impact on clinical outcomes.

**Research In Context**
Evidence before this study:
Large Language Models (LLMs) have demonstrated potential in clinical applications, particularly in tasks requiring domain-specific expertise. Previous studies explored their ability to generate accurate responses to medical queries, including preoperative assessments and management recommendations. However, these studies often relied on generic LLMs without the integration of localized clinical guidelines or secure frameworks tailored to institutional practices. No prior research has evaluated the deployment of a secure LLM system like PEACH within an operational healthcare environment. This study bridges that gap by assessing real-world application of a secure LLM model

**Added value of this study:**
This study represents the first real-world deployment and evaluation of a secure, localized LLM

system in perioperative medicine. By integrating institutional guidelines into the PErioperative AI CHatbot (PEACH), we demonstrated its ability to provide accurate, context-specific, and safe clinical recommendations. The study highlights PEACH's adaptability to protocol refinements, with improved accuracy following updates, and underscores its potential as a reliable clinician support tool.

Graphical Abstract:

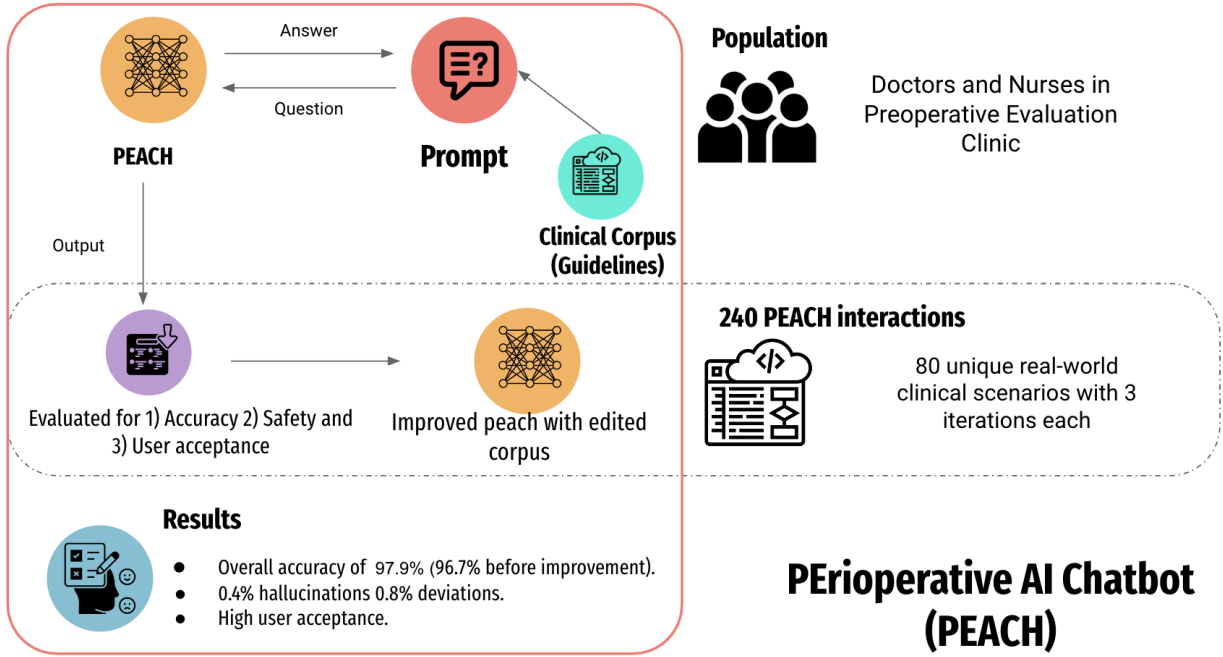

## Introduction

Large Language Models (LLMs) have demonstrated promising potential for clinical applications, particularly in tasks requiring domain-specific knowledge[1,2]. Previous research conducted by our team on open-source LLMs with RAG has shown their ability to efficiently and consistently assess a patient's fitness for surgery and provide accurate perioperative management plans[3].

Preoperative management is a complex domain that stands to benefit from advancements in LLM technology. Preoperative assessment and management errors—such as deviations from guidelines, incomplete assessments, or incorrect instructions—can result in day-of-surgery cancellations, leading to substantial economic and operational burdens[4]. Operating room expenses are estimated at USD 1400 to 1700 per hour, underscoring the importance of reducing inefficiencies[5]. By incorporating local guidelines into LLMs, these systems can offer more precise, context-aware recommendations tailored to specific institutional practices, enhancing decision-making in perioperative clinics.

However, using real-world clinical data in LLMs raises substantial challenges regarding data privacy and the potential for information leakage. For instance, adversarial attacks on LLMs have successfully extracted personally identifiable information and sensitive conversations[6]. Even anonymized patient data is vulnerable to reidentification by advanced algorithms[7].

To address data privacy challenges in deploying LLMs with real-world clinical data, we used PAIR[8] as a mitigation strategy. PAIR was developed by GovTech[9], the agency driving Singapore's Smart Nation initiative[10]. It is designed to enhance productivity while ensuring strong data security. PAIR provides a secure environment for data storage and processing. It supports data classified as "Restricted," "Sensitive," and "Normal," making it ideal for healthcare applications.

We developed the PErioperative AI CHatbot (PEACH) by integrating 35 local perioperative guidelines into the PAIR Chat environment, leveraging Claude 3.5 Sonet as its base LLM from Anthropic (San Francisco, CA, USA)[11]. Claude 3.5 Sonet, the successor to Claude 3.0 Sonet, has been reported to achieve a higher intelligence benchmark score compared to Claude-3 Opus[12].

To our knowledge, this represents the first real-world deployment of an LLM system within perioperative medicine. This study investigates the deployment and evaluation of PEACH in a clinical setting. We employed a silent co-pilot model, wherein end-users submitted inquiries to the system while continuing standard clinical management practices. The recommendations generated by PEACH were retrospectively compared against real-world management decisions.

The primary objective of this study is to evaluate PEACH's accuracy in preoperative decision-making. Secondary objectives include: (1) assessing the frequency of hallucinations (recommendations that could result in patient harm) and deviations (minor errors causing minimal harm), (2) gauging end-user perceptions of ease of use and usefulness through the Technology Acceptance Model (TAM), and (3) determining the level of agreement between

PEACH's recommendations and senior clinicians' decisions. The study seeks to provide critical insights into the clinical applicability, safety, and operational efficiency of PEACH in our dynamic healthcare environments.

## Methods

### Large Language Model Pipeline

PEACH leveraged the large context limits of advanced language models to directly input structured text data into the model, rather than employing traditional RAG techniques[13]. While RAG methods rely on chunking documents and retrieving relevant portions using search algorithms, PEACH circumvented these steps due to the enhanced performance of newer models with expanded context windows. Instead, structured input data, coupled with prompt engineering, ensured accurate and efficient integration into the language model's reasoning process. The embedding model utilized was Claude-Sonet 3.5, optimized for handling long-form contextual inputs. Retrieval parameters, such as chunk size and similarity thresholds, were therefore not relevant, as the entire structured text was directly input into the model.

To ensure the accuracy and relevance of the outputs, multiple rounds of internal testing and user validation were conducted. Iterative prompt refinement reduced hallucinations and improved the accuracy of responses. Continuous user feedback informed adjustments to the prompt engineering process, ensuring that the Chatbot reliably produced clinically relevant outputs.

### Prompt Engineering

Prompt engineering was guided by the principles outlined by Bertalan et al.[14], emphasizing specificity and contextualization to ensure comprehensive and clinically relevant responses from the LLM. "Role-playing" prompts were structured to explicitly instruct the LLM to assume the perspective of an expert preoperative clinician. The development process was iterative, involving repeated cycles of prompt design, response generation, and evaluation. Adjustments were made to improve the clarity of instructions, incorporate additional context, and minimize ambiguities. The final engineered prompt of PEACH is found within Supplementary Material, Table 1.

### PEACH interface

The PEACH interface is designed for clinical utility. It is accessible exclusively through hospital-issued, encrypted laptops, and is restricted to healthcare providers with authorized hospital-issued email accounts only. The login requires multi-factor authentication. The interface allows users to access patient data through electronic health records (EHRs), and users can copy and paste relevant patient information into the chatbot for analysis.

PEACH provides five core functions tailored to perioperative care: 1) Answers perioperative-related anesthesia questions, 2) Assists in drafting memos to other healthcare disciplines, 3) Evaluates patient suitability for telephone screening 4) Summarizes key anesthesia-related issues from the patient's medical history and 5) generates comprehensive, guideline-based perioperative management plans, including fasting instructions, suitability for carbohydrate loading, medication guidance, and additional recommendations, such as ensuring valid blood cross-matching.

To ensure responsible use, every interaction with the chatbot is accompanied by disclaimers. These disclaimers emphasize the AI-generated nature of the responses, clearly stating that they are intended as a supplementary tool and cannot replace direct clinical judgment (Supplementary Material, Table 1).

## Evaluation Framework
### A. Preoperative Guidelines
This study 35 local protocols from a major tertiary hospital in Singapore, adapted from established international perioperative standards (Supplementary Table 2). All guidelines were reviewed and were summarized into text as the PAIR chatbot has a maximum word count limit of context stuffing (45,000 words). PEACH utiltized 35,634 words (79% of context limit used) after all the protocols were uploaded.

### Usual clinical care in Preoperative Evaluation Clinic (PEC)
The PEC at our tertiary hospital manages around 120 patients daily with a team of nurses, junior anesthesiologists, and a senior attending physician. Standardized care is supported by 35 comprehensive local perioperative guidelines with more than 400 pages, and is accessible on the hospital intranet. Patients are assessed through telephone screening by nurses if criteria are met, or in-person by junior doctors. When uncertainties arise, staff refer to the guidelines or seek advice from the attending physician.

### Silent deployment
A silent deployment model was implemented with a team of five nurses, each with over two years of experience in the preoperative clinic, and five junior doctors with at least six months of anesthesiology clinical practice. Participants were briefed on the chatbot's functionality and limitations before beginning the study. They were instructed to carry out their usual clinical duties as normal, including making decisions based on their clinical judgment or consulting with the attending perioperative physician when necessary. Only after finalizing their clinical decisions were they allowed to input their questions into the chatbot and access its answers. Participants then indicated whether the chatbot's recommendations aligned with those made in consultation with the attending physician, if relevant. Figure 1 illustrates the workflow of the silent deployment process. Ethics approval was reviewed and determined to be unnecessary by the Research Quality Office of Singapore General Hospital, as the study was classified as a quality assessment initiative.

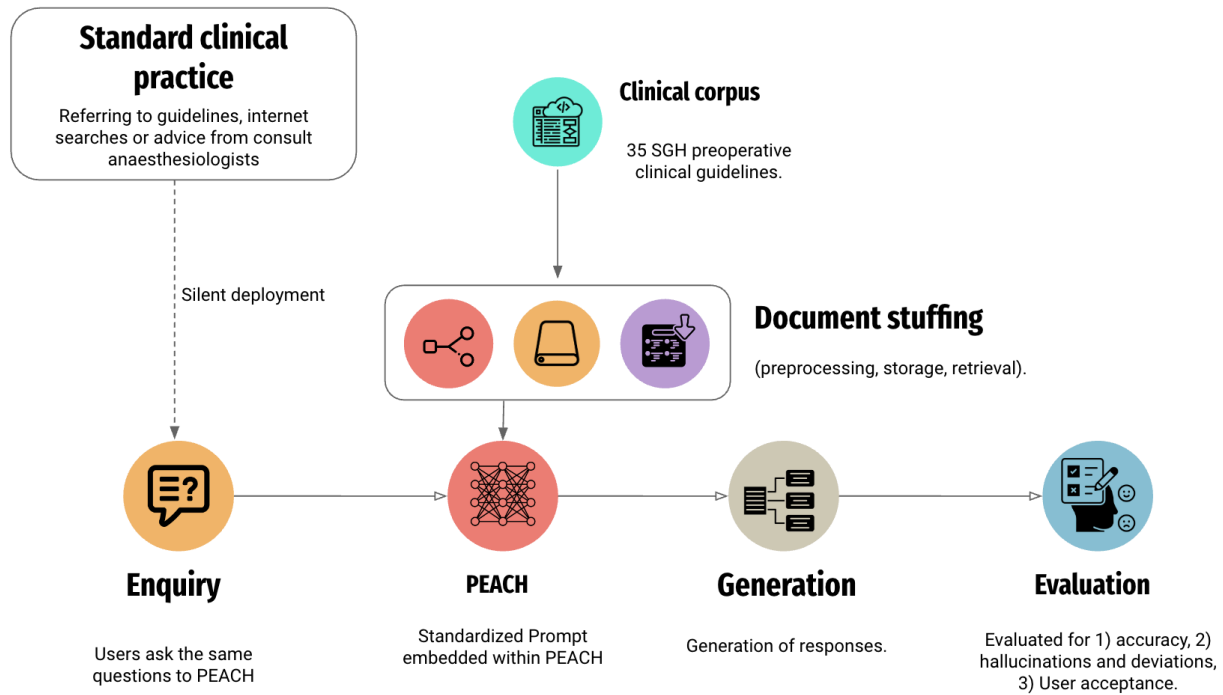

Figure 1: Silent deployment framework of PEACH.

**Evaluation Framework**

The evaluation of PEACH was designed to assess three critical dimensions: accuracy, safety, and user acceptability. The grading process for accuracy and safety was intentionally designed to be stringent and rigorous to ensure that the evaluation of PEACH's outputs was thorough and uncompromising. User acceptability was measured using the established framework of the Technology Acceptance Model (TAM) to gauge the system's practical usability and adoption by clinicians.

**Primary Outcome: Accuracy**

The primary outcome of the evaluation was the accuracy of the recommendations provided by PEACH. Given the nuanced and context-specific nature of perioperative decisions where evidence-based guidelines may not exist for all scenarios, a tiered approach was employed to determine the accuracy of recommendations. Institutional guidelines served as the first standard of reference, offering a comprehensive framework for most scenarios (e.g., stopping ACE inhibitors/ARBs before surgery unless the patient has poorly controlled hypertension). For questions not addressed by institutional guidelines, international guidelines and the current body of evidence, including systematic reviews, were used.

In cases where neither institutional nor international guidelines provided specific answers, expert consensus was sought. Accurate recommendations in these instances were determined through group discussions among experts, allowing for multiple answers as long as they were safe and clinically reasonable (e.g., deciding whether an ASA 3 patient undergoing a minor procedure like a colonoscopy should be admitted the same day or treated as day surgery).

Recommendations were deemed inaccurate if the chatbot provided advice despite lacking sufficient contextual information or when the output was inconsistent with the available evidence or guidelines. The hypothesized accuracy of PEACH was 95.0%. Binomial test were conducted to compare the accuracy of the output.

**Secondary Outcome: Deviations and Hallucinations**
In addition to accuracy, PEACH's outputs were analyzed for deviations and hallucinations. Deviations were defined as outputs that diverged from institutional guidelines or standard clinical practice but did not harm patients. For example, an output might omit mentioning the need for a cervical spine X-ray as a reason for unsuitability for TPS but still correctly conclude that the patient is unsuitable. Hallucinations, on the other hand, represented significant deviations with the potential to cause patient harm. These included instances where the AI: (1) provided information contradicting institutional guidelines, (2) suggested management plans that were dangerous or could compromise care, or (3) addressed irrelevant details not included in the clinical scenario.

A total of five independent anesthesiologists participated in grading the chatbot's performance. Two junior anesthesiologists and one attending anesthesiologist evaluated the accuracy, deviations, and hallucinations of PEACH's outputs.

**Consistency and Reliability**
The consistency of PEACH's outputs was evaluated by submitting the same set of questions to the chatbot twice more retrospectively and comparing the results across all three iterations. To evaluate inter-rater reliability (IRR), two additional attending anesthesiologists provided their clinical decisions for the same set of questions, alongside the original attending anesthesiologist who made decisions during the silent deployment. Cohen's Kappa analysis was performed to measure the agreement between the decisions of the attendings and the outputs generated by PEACH, as well as between the attendings themselves.

Additionally, distinct-n-grams analysis was conducted to evaluate the diversity of PEACH's responses across the three iterations. This analysis quantified variations in the linguistic and semantic structure of the outputs, providing insights into how consistently the chatbot formulated its recommendations.

**PEACH refinements**

Following the initial silent deployment, PEACH's outputs were systematically reviewed to identify omissions or gaps within the perioperative protocols. Based on this analysis, the perioperative guidelines uploaded to PEACH were updated to address the identified deficiencies. The revised outputs generated by the updated PEACH system were subsequently evaluated across three iterations. These refined outputs are referred to in the results section as "Updated PEACH."

**Technology Acceptance Model (TAM) Evaluation**

Two junior doctors evaluated the usability of PEACH: one who had participated in the silent co-pilot phase and another who had not been involved. The assessment focused on key aspects, including the ease of understanding PEACH's responses, the perceived safety and reliability of its recommendations, the objectivity and evidence-based nature of its outputs, and the explainability of its reasoning. These factors were graded on a Likert scale ranging from 1 (least agreement) to 5 (strongest agreement). Additionally, the junior doctors were asked whether PEACH facilitated faster clinical decision-making, with response options of "Yes," "No," or "Neutral".

All statistical evaluations are performed in the Excel and Python 3.6 environment.

**Sample size calculation**
A minimum sample size of 73 cases was determined to be necessary to estimate a 95% agreement with a 95% confidence interval and a 5% margin of error. A total of 80 PEACH interactions were targeted to account for potential invalid questions or interactions.

## Results

A total of 240 PEACH interactions, representing 80 distinct clinical questions, were analyzed. PEACH generated results within 10–15 seconds on average. The queries spanned various real-world clinical scenarios for patients undergoing elective surgeries and were categorized into general questions (46/80, 57.5%), identification of perioperative issues (22/80, 27.5%), and perioperative management queries (12/80, 15.0%). Doctors and nurses contributed nearly equal proportions of queries, with 39 queries (48.8%) from doctors and 41 (51.2%) from nurses. Among the 69 scenarios that included a complete patient history, 42 cases (60.9%) involved ASA 2 patients, while 27 cases (39.1%) involved ASA 3 patients. This reflects the typical demographics of our elective surgical population. In two instances where users entered only a generic header without patient details, PEACH requested additional information rather than generating hallucinated outputs; these interactions were included in the evaluation. An example of the clinical scenario and the PEACH output can be found in Table 1.

**Accuracy and Safety**

PEACH demonstrated an accuracy of 97.5% (78/80) in the first iteration of responses generated by users. Across the subsequent two iterations, the overall accuracy was 96.8% (232/240). PEACH produced accurate answers across all three iterations for 75 out of 80 questions (93.8%). Hallucination rates were minimal, occurring in 4 out of 240 outputs (1.7%), while deviations were found in 2 out of 240 outputs (0.8%) (Table 2).

The initial outputs from PEACH aligned with the attending anesthesiologist's recommendations in 53 out of 58 cases (91.4%). There was only one scenario where all 3 iterations had inaccurate answers - PEACH instructed to stop empagliflozin 1 day before the surgery instead of 3 days before. Table 3 summarizes all the hallucinations and deviations present from the PEACH outputs.

**Updated PEACH**

After reviewing the protocols uploaded on SGLT2 inhibitors, we noticed that the protocol says "SGLT2 inhibitors (eg. dapagliflozin): To stop for 3 days". The protocol was subsequently amended to "SGLT2 inhibitors (eg. dapagliflozin, empagliflozin and all other drugs in the same family): To stop for 3 days". The scenario was replicated, and all 3 iterations were correct and able to highlight the risk of euglycemic ketoacidosis. This would bring an overall accuracy of 97.9% (232 / 240) with a hallucination rate of 1 out of 240 outputs (0.4%). This is statistically significant compared to the null hypothesis of 95% accuracy (p=0.018, 95% CI (0.952, 0.991)). Supplementary Material Table 2 further details the old and new responses with a change of wording within the protocol.

**User acceptability**

User perceptions of PEACH were positive. The chatbot was rated highly for ease of understanding (mean 4.32 (SD 0.53)), safety (mean 4.7 (SD 0.79)), objectivity and non-bias (mean 4.75 (SD 0.54)), and explainability (mean 4.46 (SD 0.62)). Importantly, the users identified that 95.0% (152/160) of the time PEACH's responses would help them make decisions better or faster, with only 5.0% (8/160) responding negatively.

Among the PEACH iterations, substantial agreement was observed (Kappa 0.772 and 0.893 between the 3 iterations). Among the attending consultants, moderate agreement was observed in pairwise comparisons (0.610 and 0.784). This variability reflects differences in individual clinical judgment, underscoring the potential of PEACH to provide consistent support across diverse scenarios (Figure 2).

The average distinct n-gram scores across all clinical scenarios were 0.309 for 1-gram and 0.592 for 2-gram, reflecting a moderate level of linguistic diversity in word usage and phrase construction (Figure 3). Scores within the range of 0.3 to 0.6 suggest moderate diversity, which is ideal for a clinical chatbot. This balance ensures that the chatbot provides some variation in responses to avoid rigidity while maintaining consistency and adherence to clinical standards.

**Table 1:** An example of PEACH enquiry and answer of an accurate. The sensitive information within the scenario is redacted with (x) for privacy reasons.

| Enquiry by User |
|---|
| "81YO Chinese Lady<br>ADLi comm ambulant<br>Lives with husband<br>NKDA<br>Social drinker, non smoker<br><br>===Renal history===<br>1. ESRF secondary to DKD on HD 2/4/6 via left RC AVF<br>- Left RC AVF created by xxx on xx/xx/22<br><br>a) Anemia with hx of LBGIT 2023 |

- OGD (xx/x/23): 2 large pedunculated polyps in lesser curve; 1 of which surface slightly ulcerated w stigmata of recent bleed. Left alone as not bleeding
- OGD (xx/x/23): No blood in stomach or duodenum. 2 x pedunculated polyps in stomach body. Not actively bleeding. Smaller polyp taken with hot snare polypectomy, 2 x Boston Resolution clips to base. Large polyp Endoloop applied to base, hot snare polypectomy above Endoloop, 2 x Boston Resolution clips + 1 x Instinct clip to base. No ulcers
- Colonoscopy (xx/x/23): Capacious colon. Ascending colon diverticulosis. Not actively bleeding

b) Recurrent intra-dialytic hypotension and syncopal eps; b/g supine/ rest hypertension
- Holter (across SLED) (x/x/24): Baseline sinus rhythm with first degree AV block, infrequent supraventricular ectopics (<1% burden), highest grade 3 beats non-sustained atrial tachyardia at 106 bpm.
- EEG (x/x/24): This EEG recorded evidence of electrophysiologic dysfunction in the left temporal region, on a background of a mild and diffuse encephalopathy.
- ANS (xx/x/24): suggestive of sudomotor and cardiovascular dyutonomia.
- TTE (xxx 2024) : Normal LV systolic function. LVEF 60 % (LVEF Simpson biplane). No regional wall motion abnormality. Tiny pericardial effusion.
- CVM review (xx/x/24):
 > Recent Holter done during dialysis did not show any significant arrhythmias.
 > No gross abnormalities on recent echo to explain hypotension and syncope as well.
 > Previously had LV intra-cavity gradient, but this was no longer seen in latest TTE
- 8am cortisol 282 - SST 413 > 531 > 551 (xxx 2024)
- Started midodrine 5mg 3x/week pre-dialysis in xxx 2024 > stopped > restarted at 2.5mg in xxx 2024

=== Past Medical History ===
1. PAF (CHADSVASc 6)
- OAC held off due to high bleeding risk/recurrent BGIT and ESRF
- MIBI x/x/22 Negative for ischaemia
- TTE xxx 2023: LVEF 60-65% no RWMA, severe LA dilatation, no pericardial effusion or thrombus
2. HTN
3. DM
4. LBGIT sec AVM (xxx and xxx 2023) and AVM in the terminal iluem
5. Incidental findings of left small renal mass ?clear cell RCC seen on CTMA on xx/xx/23
- Last seen by Uro on xx/x/24 -- planned for US surveillance IVO age and premorbids

Last admitted xxx 2024 for Hypertensive urgency
- Started Valsartan 40mg BD on dialysis days, Valsartan 80mg BD on non-dialysis days, nifedipine 30mg OM
- Midodrine 2.5mg 3x/week on dialysis days, serve 30 min before

===HOPC===
noted NKF memo - intradialysis noted LOC < 10s, noted drop of BP 93/38, completed dialysis of 2hours and 15min, after returning all bloods, patient regained consciousness, BSL 12.2,
History from patient
pre-syncope no symptoms, no cp sob giddiness palpitation diaphoresis
has recurrent intradialytic syncope ? last attack a few months ago
after wokeup cannot rmb the event, and no sleepiness

| |
|---|
| recently well no fever cough ARI diahrrea or nausea/vomiting
no headache weakness numbness

Does this patient require a cardiac referral?" |
| PEACH Full Response |
| Based on the information provided, this patient **does not require a new cardiac referral at this time**. Here's the rationale:

Recent cardiac evaluations:
- The patient has had recent cardiac assessments, including: a) Holter monitoring (x/x/24) during dialysis b) Transthoracic echocardiogram (TTE) in xxx 2024 c) Cardiology review on xx/x/24

Cardiac status:
- Normal left ventricular systolic function (LVEF 60-65%)
- No regional wall motion abnormalities
- Previous LV intra-cavity gradient is no longer present
- Tiny pericardial effusion noted, which is not clinically significant

Arrhythmia:
- Known paroxysmal atrial fibrillation (PAF)
- Recent Holter showed no significant arrhythmias during dialysis

Ischemic evaluation:
- MIBI on x/x/22 was negative for ischemia

Current management:
- The patient is already on appropriate medications for blood pressure control
- Midodrine is being used to manage intradialytic hypotension

Recent cardiology input:
- The patient was reviewed by cardiology on xx/x/24, where it was noted that there were no gross abnormalities to explain the hypotension and syncope

However, there are some important points to consider:
- The patient has recurrent intradialytic hypotension and syncopal episodes, which are likely related to her ESRF and autonomic dysfunction rather than primary cardiac issues.

The patient's cardiovascular status should be closely monitored perioperatively due to the history of PAF and risk of hemodynamic instability.

If there are any new cardiac symptoms or significant changes in the patient's condition before surgery, a repeat cardiology consultation may be warranted.

The anesthesia team should be prepared to manage potential cardiac complications, including arrhythmias and blood pressure fluctuations, during the perioperative period.

In conclusion, based on the recent cardiac evaluations and stable cardiac status, a new cardiac referral is not necessary at this time. However, close communication with the nephrology team and careful perioperative cardiac monitoring are essential. |

```
Protocols for reference:

Preoperative cardiac referral guideline and workflow.docx
Perioperative Management of patients with hypertension.docx
==============================
Thank you for using PEACH.
To view full guidelines, go to [SGH Intranet > Div of Anesthesia and Perioperative Medicine > PAC guidelines].

Disclaimer: This response is generated by AI and is not a substitute for clinicians. Users are responsible for all decisions made based on this information. Use with caution.
```

**Table 2:** Characteristics of clinical questions and the accuracy and safety of PEACH's responses. The "Updated PEACH" outputs were evaluated following revisions to the protocol, specifically addressing the management of SGLT2 inhibitor medications.

| | | | Count (%) / Mean (SD) | Updated PEACH |
|---|---|---|---|---|
| Clinical Scenario | Questions | | 46 / 80 (57.5%) | |
| | Issues | | 22 / 80 (27.5%) | |
| | Management | | 12 / 80 (15.0%) | |
| ASA | NA | | 11 | |
| | | 2 | 42 / 69 (60.9%) | |
| | | 3 | 27 / 69 (39.1%) | |
| Asked by | Doctors | | 39 / 80 (48.8%) | |
| | Nurses | | 41 / 80 (51.2%) | |
| Accuracy | First generation accuracy | | 78 / 80 (97.5%) | |
| | All 3 iterations are correct | | 75 / 80 (93.8%) | 76 / 80 (95.0%) |
| | Overall accuracy | | 232 / 240 (96.7%) | 235 / 240 (97.9%)* |
| Agreeable with primary consultant | NA | | 22 | |
| | Yes | | 53 / 58 (91.4%) | |
| | No | | 5 / 58 (8.6%) | |
| Safety | Hallucinations | | 4 / 240 (1.7%) | 1 / 240 (0.4%) |
| | Deviations | | 2 / 240 (0.8%) | 2 / 240 (0.8%) |
| | Disclaimer/warning | | 240 / 240 (100.0%) | 240 / 240 (100.0%) |
| TAM (1-5) | Ease of understanding | | 4.32 (0.53) | |

|  | Safety | 4.7 (0.79) |  |
| --- | --- | --- | --- |
|  | Objectivity, non-biased against any condition or demographics | 4.75 (0.54) |  |
|  | Explainability (good justifications provided) | 4.46 (0.62) |  |
| Will this response help you to make decision better/faster | Yes / Neutral | 152 /160 (95.0%) |  |
|  | No | 8 / 160 (5.0%) |  |
| Answer found within guidelines | NA | 26 |  |
|  | Yes | 27 / 54 (50.0%) |  |
|  | No | 27 / 54 (50.0%) |  |

*p-value: 0.018; 95% CI: (0.952, 0.991); Binomial test conducted with a null hypothesis of 95% accuracy.

**Table 3:** Summary of scenarios with hallucinations and deviations from PEACH.

| Scenario Summary | Wrong PEACH responses | Number of wrong responses | Deemed level of harm |
| --- | --- | --- | --- |
| Hallucinations | | | |
| ASA 3, 69-year-old patient with a history of diabetes on empagliflozin, dilated cardiomyopathy, morbid obesity (BMI 52.6), asthma, and thyroid nodule, presenting for cataract surgery. User asked for perioperative management. | Stopping empagliflozin for 1 day instead of 3 days[15]. | 3 | Major |
| If methotrexate (MXT) should be stopped preoperatively in a patient with well-controlled Rheumatoid Arthritis. | Stating that MXT should be stopped, despite current international consensus indicating that it is unnecessary[16]. | 2 | Moderate |
| If a patient with an unclear cardiac history (Previously seen by cardiology for active cardiac symptoms, lost to follow-up since, but now claims to have good effort tolerance) can be seen by telephone screening (TPS). | Patient can be seen by TPS. Answer deemed to be unsafe without further clarification of the history. | 1 | Major |
| A patient scheduled for excision of a right buccal mass, with a history of hypertension, diet-controlled hyperlipidemia, and mildly elevated creatinine, as well as past emergency visits in 2021 and 2024 for hypertensive urgency on 3 antihypertensive agents, but is now "well". Asking for suitability for TPS. | Stating patient is fit for TPS, but based on our guidelines, the patient would be considered to have "uncontrolled" hypertension and should be reviewed in person. | 1 | Moderate |
| User only gave information about a 59-year-old patient with Hypertension, and a previous ultrasound liver showing fibrofatty changes. Asked if the patient requires preoperative electrocardiography (ECG). | Instead of stating there was insufficient information, gave the suggestion of doing ECG. | 1 | Minor |
| Deviations | | | |
| ASA 3, 69-year-old patient with a history of benign prostatic hypertrophy, type 2 diabetes with macular retinopathy, hypertension, hyperlipidemia and chronic kidney disease scheduled for MRI prostate biopsy. There was a previous Cervical spine X-ray which showed mild loss of C5/6 and C6/7 space. User asked if the patient is suitable for TPS. | Did not mention the cervical spine X-ray as a consideration for why the patient is not fit for TPS. | 1 | Minimal |
| Asking if a patient undergoing examination under anaesthesia for rectal cancer after previous ultra low anterior resection (ULAR) can undergo spinal anesthesia for the procedure. | Mentioned that ULAR may have "altered local anatomy, making spinal anesthesia more difficult and less effective" as part of the recommendation to suggest general anesthesia (GA) for this patient (But correctly suggested that spinal anesthesia is not contraindicated, although GA would be preferred in this case). | 1 | Minimal |

**Figure 2:** Inter-rater reliability (Cohen's Kappa) between PEACH iterations and attending consultants. The diagonal values demonstrate perfect agreement within individual decision sets (Kappa = 1).

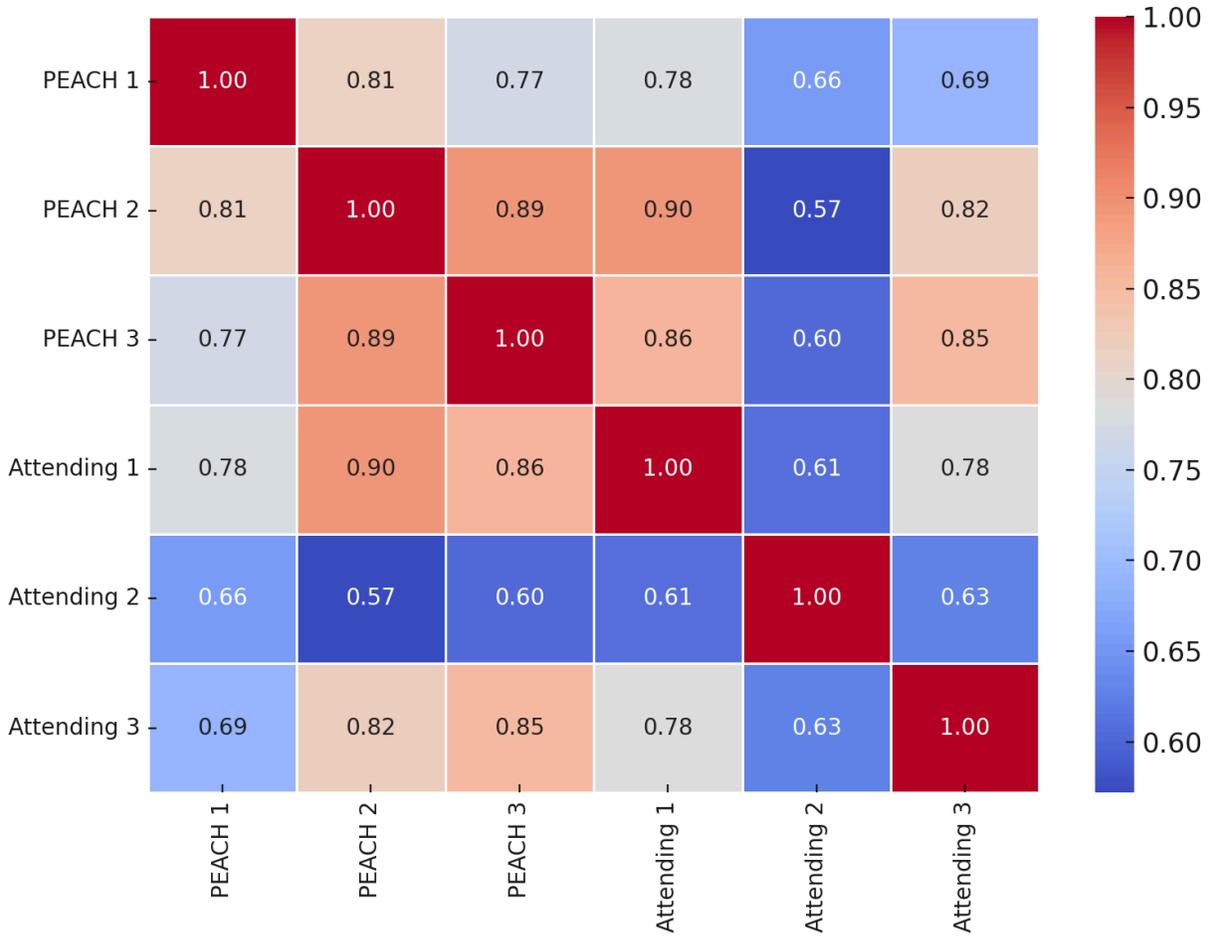

**Figure 3:** The distinct 1-gram and 2-gram across the 80 clinical scenarios when comparing 3 iterations of PEACH.

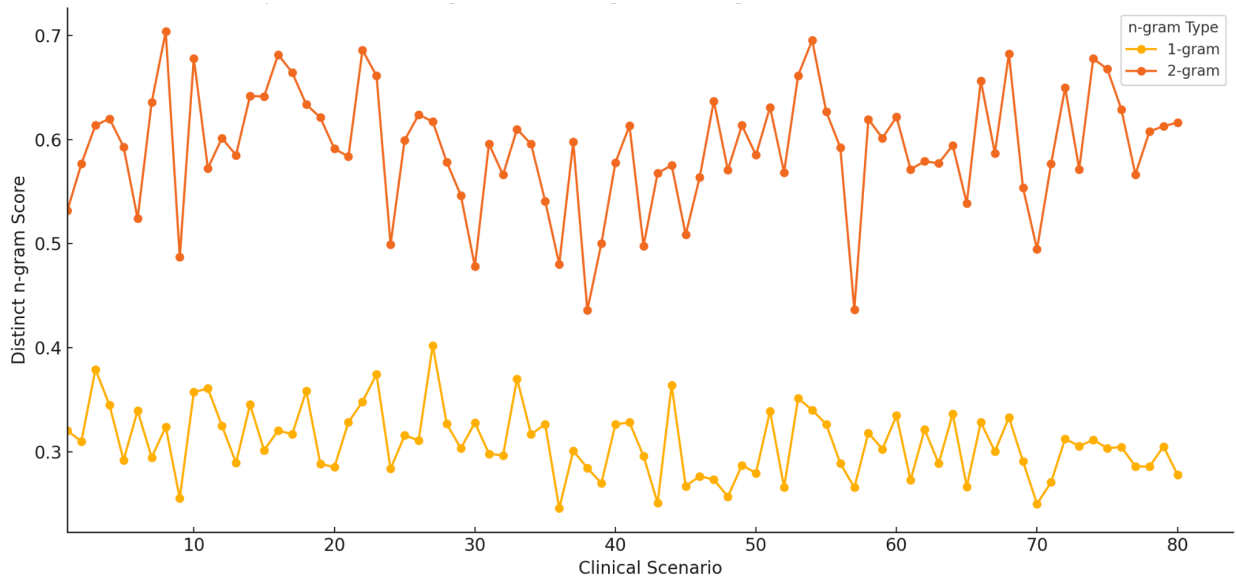

## Discussion

This study underscores the feasibility and utility of integrating a secure LLM framework into perioperative clinical workflows. PEACH demonstrated high levels of accuracy, safety, and user acceptability, establishing itself as a valuable clinical adjunct in preoperative medicine. The findings emphasize the potential of LLM systems to reduce variability in decision-making and enhance operational efficiency in perioperative settings.

The adaptability of PEACH was also evident through its iterative refinement process. The significant improvement observed after updating the protocol for SGLT2 inhibitor management highlights the model's capacity to evolve alongside clinical practice. This adaptability ensures that PEACH remains aligned with institutional guidelines and emerging evidence, reinforcing its role as an amendable and robust tool. Such flexibility is critical in dynamic healthcare environments, where clinical protocols are frequently updated to reflect new evidence and practices.

### Standardizing Preoperative Care and Reducing Variability

One of the key challenges in perioperative medicine is the inherent subjectivity in clinical decision-making, influenced by variations in individual judgment and risk tolerance[17]. Tools like PEACH address this challenge by promoting consistency and standardization. This consistency is particularly advantageous in perioperative care, where standardized evaluations can reduce miscommunication and prevent conflicts within the care team. Moreover, a consistent approach helps coordinate decisions with surgeons and ensures equitable patient care. Misaligned evaluations could unnecessarily delay surgery, disrupt workflows, and increase costs (E.g. anesthesiologist cancelling the surgery on the day of operation, when the case was already cleared by a previous preoperative attending).

The integration of PEACH into preoperative workflows holds the promise of improving efficiency in the broader perioperative landscape. By providing consistent, guideline-based recommendations, PEACH can streamline preoperative processes and reduce the operational burden on healthcare providers.

### Enhancing Workflow Efficiency with Human Oversight

A valuable application of PEACH lies in augmenting preoperative workflows through a "human-in-the-loop" framework. In many preoperative clinics, initial patient screening determines whether evaluation by a nurse or doctor is required, or if a low-risk patient can be seen on the day of surgery. PEACH's ability to triage such decisions safely and efficiently has the potential to save significant time and reduce costs. Additionally, by assisting clinicians in drafting patient instructions, PEACH could alleviate administrative burdens and help reduce clinician burnout. Importantly, the model operates as a support tool rather than an autonomous decision-maker, ensuring that all recommendations are reviewed by qualified clinicians. This framework balances efficiency with accountability, with healthcare providers maintaining oversight and responsibility for patient care.

The results of this study are particularly relevant within the evolving landscape of elective surgical services, where reduced hospital stays and outpatient preoperative assessments have become the norm[18]. By providing rapid, context-aware recommendations, PEACH supports this transition and enhances the efficiency of perioperative workflows. Its capability to process vast amounts of data and generate responses based on comprehensive, updated guidelines positions it as a valuable tool in standardizing preoperative assessments.

## Limitations and Future Directions

This study offers valuable insights into the integration of LLMs like PEACH into perioperative workflows but has several limitations. First, the protocols embedded within PEACH were specific to a single tertiary hospital. As clinical guidelines and practices vary across institutions and regions, the findings may not be fully generalizable. Adapting PEACH to other settings would require careful customization of its protocol database to reflect local practices.

Another limitation is the dependence of PEACH's accuracy on the quality and completeness of the embedded guidelines. Errors or omissions in the protocols, such as the initial mismanagement of SGLT2 inhibitors, highlight the importance of ongoing updates and rigorous review of clinical content. Additionally, the TAM evaluation involved a small number of junior doctors, which may not fully represent the perspectives of all healthcare providers, including senior clinicians.

Finally, while PEACH demonstrated high accuracy and safety in its responses, there remains a risk of over-reliance on AI by users. Clinicians must remain vigilant and use PEACH as a supplementary tool, not a replacement for clinical judgment. Future studies should investigate the impact of such systems on clinical outcomes and explore scalability in diverse healthcare environments.

Future research should explore the scalability of PEACH and its adaptability across different clinical specialties. Expanding its application beyond perioperative medicine to fields such as surgery of internal medicine could further validate its utility. Additionally, investigating the integration of advanced retrieval-augmented generation (RAG) frameworks, such as LlamaIndex, could enhance PEACH's ability to retrieve and synthesize information from increasingly complex datasets.

## Conclusion

PEACH has demonstrated high accuracy, consistency, and strong user acceptance, with minimal rates of hallucinations. Its adaptability and ease of refinement further highlight its potential as a valuable tool in clinical workflows. By improving decision-making consistency and enhancing operational efficiency, PEACH shows promise as a reliable clinician support system in perioperative medicine. Future research should explore the scalability and applicability of LLM models like PEACH across diverse medical specialties and clinical settings.

## Declaration Statements

### 1. Data Availability:
All data used in this study are available upon request and have been de-identified in compliance with applicable privacy laws and institutional policies.

### 2. Code Availability:
The code for the project can be found at https://github.com/yuheke36/PEACH.git.

### 3. Acknowledgements/ Funding:
The authors sincerely thank the nurses at Preoperative evaluation clinic as well as Dr. Roderica Ng, Dr. Jocelyn Gao, Dr. Lam Hsin Hui and Dr. Tan Chun Lei from Singapore General Hospital for their invaluable contributions as human evaluators in this study. Their dedication and insightful input enriched the research. We also acknowledge the preoperative clinical guidelines provided by the hospital, which were pivotal in the successful execution of this project. The project is not funding by any funding agency in the public, commercial, or not-for-profit sectors.

### 4. Author Contributions:
Conceptualization and guidance: YH Ke, HR Abdullah, KWT Sin, DSW Ting
Product development: BWX Ong, CY Oh
Coding and technical development: K Elangovan, L Jin, JMH Cheng, AHY Lee
Data analysis: YH Ke, L Jin
Manuscript preparation and proof-reading: All authors

### 5. Competing Interests:
The authors declare no competing interests.
### References:

1. Thirunavukarasu AJ, Ting DSJ, Elangovan K, Gutierrez L, Tan TF, Ting DSW. Large language models in medicine. *Nat Med*. 2023;29(8):1930-1940. doi:10.1038/s41591-023-02448-8

2. Ke Y, Jin L, Elangovan K, et al. Development and testing of retrieval Augmented Generation in large Language Models -- A case study report. *arXiv [csCL]*. Published online January 29, 2024. https://scholar.google.com/citations?view_op=view_citation&hl=en&user=uuI-bysAAAAJ&citation_for_view=uuI-bysAAAAJ:Zph67rFs4hoC

3. Ke YH, Jin L, Elangovan K, et al. ORetrieval Augmented Generation for 10 large Language Models and its generalizability in assessing medical fitness. *arXiv [csCL]*. Published online October 10, 2024. http://arxiv.org/abs/2410.08431